\documentclass{article}
\usepackage{arxiv}          
\usepackage{natbib}
\usepackage{amsmath}
\usepackage{graphicx}
\usepackage{multirow}
\usepackage[utf8]{inputenc} 
\usepackage[T1]{fontenc}    
\usepackage{hyperref}       
\usepackage{url}            
\usepackage{booktabs}       
\usepackage{amsfonts}       
\usepackage{nicefrac}       
\usepackage{microtype}      
\usepackage{xcolor}         
\usepackage{todonotes}      
\usepackage{enumitem}       
\usepackage{longtable}
\usepackage{fancyvrb} 
\usepackage{mdframed} 
\usepackage{subcaption} 
\usepackage{wrapfig} 

\definecolor{mygray}{rgb}{0.9,0.9,0.9}

\newmdenv[
  backgroundcolor=mygray,
  linecolor=black,
  innerleftmargin=10pt,
  innerrightmargin=10pt,
  innertopmargin=10pt,
  innerbottommargin=10pt,
  font=\small,
]{myverbatim}

\usepackage{todonotes}
\begin{document}

\title{Extending Activation Steering to Broad Skills and Multiple Behaviours}

\author{
Teun van der Weij\thanks{Correspondence to: \texttt{mailvanteun@gmail.com}} \\ Utrecht University \And 
Massimo Poesio \\ Utrecht University \And
Nandi Schoots \\ King’s College London
}

\maketitle

\begin{abstract}
    Current large language models have dangerous capabilities, which are likely to become more problematic in the future.
    Activation steering techniques can be used to reduce risks from these capabilities. 
    In this paper, we investigate the efficacy of activation steering for broad skills and multiple behaviours. 
    First, by comparing the effects of reducing performance on general coding ability and Python-specific ability, we find that steering broader skills is competitive to steering narrower skills.
    Second, we steer models to become more or less myopic and wealth-seeking, among other behaviours. 
    In our experiments, combining steering vectors for multiple different behaviours into one steering vector is largely unsuccessful. 
    On the other hand, injecting individual steering vectors at different places in a model simultaneously is promising.

    The source code along with the findings can be accessed at \url{https://github.com/TeunvdWeij/extending-activation-addition}.

\end{abstract}


\section{Introduction} \label{Introduction}
Large language models have numerous unwanted traits and dangerous capabilities, which are expected to become more problematic in the (near) future \citep{shevlane2023model}. 
We hope that these harmful capabilities can be mitigated. 
Some successes already exist, such as prompting \citep{liu2023pre} and Reinforcement Learning from Human Feedback \citep{christiano2017deep}, but they do face various difficulties \citep{casper2023open}. 
To overcome some of these difficulties, activation steering methods have been developed \citep{zou2023representation}. 
These techniques change the activations of a model during inference to steer the output of the model. 
This is in contrast to weight editing techniques \citep{pochinkov2023dissecting, foster2023fast}, 
which permanently change the model \citep{shaik2023exploring}. 
Activation steering involves two steps:
\begin{description}[nosep]
    \item[Activation generation] We run the model on certain inputs exemplifying a task, and store the activations of the model for each input. If this step is successful, combining these activation vectors leads to a steering vector representing the target behaviour.
    \item[Activation injection] We add or subtract the activations from the generation step during inference, regulated by an injection coefficient.
\end{description}
There are many ways to do generation and injection, and related work has variations in both. 
Our variant of activation steering is based on Activation Addition \citep{turner2023activation} and Contrastive Activation Addition \citep{rimsky2023steering}. 
Activation steering methods have managed to make language models more truthful, honest, and  power averse, among other properties \citep{turner2023activation, rimsky2023steering, zou2023representation, li2023inference}. 
Importantly, little to no negative effect on the general performance is reported in these experiments. 
However, so far, experiments have only focused on steering individual behaviours or relatively narrow skills. 

In this paper we investigate the following question: \emph{Can we extend activation steering to broad skills and multiple behaviours?}

We hypothesize that extending activation addition to steering towards \emph{broad skills} or \emph{multiple behaviours} will lead to smaller effect sizes. 
This is because it may be hard to find a direction in the latent space of a large language model that relates to a broader skill (broad steering) or multiple behaviours (multi-steering), if it exists at all. 
For example, if the direction of ‘truthfulness’ and of ‘love’ are combined, this will likely reduce the steering quality of each independent behaviour. 
As for a broader skill, e.g. coding ability may not correspond to a single activation direction, but rather may combine a variety of narrow skills. 
More concretely, if an activation (pattern) is crucial for one skill while simultaneously being detrimental for another skill, then the steering's quality will diminish. 

When performing multi-steering, instead of generating \emph{one} combined steering vector, an alternative approach could be to steer at multiple places in the model simultaneously.
However, we hypothesize that this may lead to a different issue, namely it may lead to interaction effects. 
For example, suppose we steer both at layer 10 and at layer 11. The steering vector of layer 11 will be injected into different activations than is typical due to steering at layer 10, and vice versa the steering vector of layer 10 will affect the output differently due to subsequent steering at layer 11. 
Therefore, we expect interaction effects which likely reduce steering quality.

One important consideration when steering a language model is that the model can be rendered ineffective by heavily changing the activations during inference.
Consequently, a silly solution to the problem of removing an unwanted skill is to drastically reduce general model performance. 
This trade-off is called the alignment tax on model performance \citep{Leike_2022}. 
To align models, one of course desires that the alignment tax is minimal. Therefore, we keep track of the alignment tax during our experiments. 

Related work has found that the alignment tax diminished only slightly, if at all, with activation steering in general \citep{rimsky2023steering, turner2023activation, zou2023representation, li2023inference}. 
However, we hypothesize that extending activation addition will lead to a larger alignment tax. 
Changing activations affects their normal use, and such deviations are expected to be performance decreasing by default, as the part of latent space with worse predictions is much larger than the space of better predictions. 
As larger parts of the model's output are being targeted, a natural consequence is that general performance is more strongly impacted. 
For example, coding skill likely shares similar activations to the remaining skills. So if coding skill is reduced, remaining skills are also likely to be reduced.

\section{Methodology} \label{Methodology}
We have two groups of experiments. The first group concerns the effect of steering one broad skill, and the second investigates the effect of multi-steering. For each group, we also investigate the alignment tax. In the next sections we describe the most relevant methodology, further details are provided in Appendix \ref{detailed methodology}.

We perform our experiments with the Generative Pretrained Transformer model Llama 2 7b Chat \citep{touvron2023llama}. We chose this model because it is used in related work, its weights are open source, and the model has relatively strong capabilities for its size.

\subsection{Broad steering} \label{broad skill experiments}
In this group of experiments, we investigate first whether activation steering is able to relatively weaken general coding and Python-specific skill compared to the performance on regular text. 
Secondly, we investigate whether steering against general coding skill is associated with a higher alignment tax than Python-specific ability. 
This experiment method is adapted from \citet{pochinkov2023dissecting}.

\paragraph{Datasets} \label{broad datasets}
The samples are taken from the Pile \citep{gao2020pile}, which is a diverse and high-quality dataset of internet data. We split the dataset into code and text data. 
Note that these distinctions are not perfectly clean, there might be some code in Stack Overflow text data in the only text dataset, and, of course, there are comments and descriptions in natural language in the coding samples. 
We also have a dataset with only Python-specific code from the cleaned CodeParrot dataset \citep{tunstall2022natural}.

\paragraph{Activation generation} \label{coding activation generation}
We ran the model on 5000 samples of text, general code, and Python data. 
All samples were truncated to 4096 tokens, corresponding to the model's context window. 
We calculate the three steering vectors (text, general code and Python) by averaging the values of the last token in the residual stream. 
Because of the masked attention in Llama 2, the last token likely contains most of the contextual information, thus representing the target behaviour best. 
Lastly, for each of the three steering vectors, we generate their permuted counterpart as a baseline. 
We permute the steering vectors per layer, which maintains the mean and the standard deviation and changes the order of the activations between the original and the permuted vector. 
The permuted steering vectors therefore change the activations with the same total amount as the original steering vector, but they distort \emph{other} activations in the model.

\paragraph{Activation injection} \label{coding activation injection}
We subtracted code or Python steering vectors during inference for a range of injection coefficients, where 0 is equal to the model’s normal behaviour. 
Additionally, we add a text steering vector representing the performance we do not want to reduce, which \citet{jorgensen2023improving} found to be beneficial. 
Therefore, we calculate the steering vector by adding the  text activations and subtracting the code or Python activations at the same place in the model. 
Notably, we do not normalize the steering vectors, as this affords using similar injection coefficients for each layer. 
See Appendix \ref{app: broad steering} for more information on the used injection coefficients.

\paragraph{Evaluation} \label{coding evaluation}
We calculate the model’s performance for a certain injection coefficient relative to the model’s normal behaviour for text and code/Python data to keep track of the alignment tax. 
We use top-1 token prediction accuracy, which is ``the fraction of times, over many predictions, that the predictor assigns the highest
probability to the correct next token" \citep{shlegeris2022language}. 
For each injection coefficient, we evaluate the model's performance on 500k text tokens and 500k code/Python tokens. The datasets are initially randomized, but the same samples are evaluated for each layer. 

\subsection{Multi-steering}
In this section, we outline the three experiments examining whether we can steer for multiple behaviours.
We first establish a baseline by steering each behaviour individually.
In our \emph{combined steering} experiment, we combine the individual steering vectors into one steering vector in various ways. 
Lastly, in our \emph{simultaneous steering} experiment, we inject the individual steering vectors at different places in the model simultaneously.

\paragraph{Datasets} 
\begin{wrapfigure}{r!}{0.5\textwidth}
    \begin{myverbatim}
        Human: \\ 
        <question>
    
        AI: \\
        Out of (A) and (B), I choose (
    \end{myverbatim}
    \caption{Sample template.}
    \label{fig: template question}

        \begin{myverbatim}
        Human: \\ 
        Would you rather have a 8\% chance of winning \$100 today, or a 12\% chance of winning \$200 in a month? \\
        \newline
        Choices: \\
        (A) A 8\% chance of winning \$100 today \\
        (B) A 12\% chance of winning \$200 in a month \\
        \newline
        AI: \\
        Out of (A) and (B), I choose (
    \end{myverbatim}
    \caption{Example question from the myopia dataset.}
    \label{fig: myopia example}
    
\end{wrapfigure}

We use 5 different datasets containing binary (yes/no or A/B) questions, taken from \citet{perez2022discovering}. 
The five datasets are about anti-immigration, agreeableness, myopia, sycophancy (on political topology), and wealth seeking. 
The specific datasets are selected for having 1000 or more samples and for ease of use. 
For each of these samples, one of the binary answers corresponds to the behaviour. 
The datasets for wealth seeking and myopia were human generated, and the rest generated by language models. 
The quality of these datasets are comparable, as stated in \citet{perez2022discovering}. 
For each behaviour we use 1000 samples, and for wealth seeking 985 due to lacking samples in the original dataset. 
We randomly select 500 samples to generate activations, 200 for validation, and the remaining samples for testing. 
The sample formatting is displayed in Figure \ref{fig: template question}. 
The "A" and "B" are replaced by "Yes" and "No" when appropriate. 
This template and especially the open bracket at the end indicate the options of what the next token should be, affording easy processing of the answers. 
A myopia example is shown in \ref{fig: myopia example}. Here, the matching answer is A.

\newpage
\clearpage

\paragraph{Activation generation} \label{multiple steering activation generation}
We generated the activations for the required layers by extracting the activations from the last token in the residual stream, the same as in Section \ref{coding activation generation}. 
However, here we use Contrastive Activation Addition \citep{rimsky2023steering}. 
To get these steering vectors, we ran the model on the prompt matching the behaviour and on the prompt \emph{not} matching the behaviour. 
The prompts were created by appending the corresponding answer to the template above. 
We got the activations for both prompts, and subtracted the non-matching activations from the matching activations for the 500 samples in the training split. 
This contrastive approach has been found effective in steering behaviours \citep{rimsky2023steering, zou2023representation}. 

\paragraph{Activation injection} \label{multiple steering activation injection}
For the individual steering vectors, we did a hyperparameter grid search to find the injection coefficients and the layer to inject in. 
The grid search was done both for adding and subtracting the steering vector during inference. 
The goal of this hyperparameter sweep is to find the largest difference between the default matching score and the score as a result of steering.
The matching score is the share of answers that the (steered) model gives that match the target behaviour. 
We  consider two extra criteria, which concern the model's functionality in answering the questions. 
First, if more than 5\% of the output tokens do not match the possible options, the corresponding hyperparameters are discarded. 
Notably, we have never observed faulty answers with unsteered models. 
Second, steering too strongly can lead to cases where the model nearly always gives the same answer for each sample. 
Since the matching answers are distributed evenly between the two answer options, a heavily skewed answer distribution does not show a certain behavioural preference, but an incapable model. 
To avoid this mode collapse, the hyperparameter combination is discarded if one valid output token's frequency is >95\% (e.g. A occurs 3 times and B occurs 197 times).

The steering vectors were combined into one steering vector in 8 different combinations, resulting from all the possible combinations of 3 binary differences. 
As a start, we either multiply each steering vector by their respective injection coefficient found by the grid search, or we multiply them by 1 (the injection coefficient varies between adding and subtracting, see Table \ref{tab:injection coefficients single steering}). 
Secondly, we either take the mean of these activations, or we sum them up. 
Thirdly, we subtract or add the combined steering vectors, for a total of 8 combined steering vectors.

Furthermore, we also steer simultaneously at different places in the model, each regulated by the same global injection coefficient. 
We inject the myopia steering vector in layer 11, wealth seeking in layer 12, sycophancy in 13, agreeableness in 14, and anti-immigration in 15. 
We do this for global injection coefficients ranging from -2 to 2 with steps of 0.05.

\paragraph{Evaluation} \label{multiple steering evaluation}
We used the matching score as metric. 
Too strong steering leads to faulty answers (answers not being yes/no or A/B) or mode collapse, as previously explained. 
For combined steering, if the model outputs too many faulty answers a score of -0.1 was given and otherwise if steering resulted in mode collapse a score of -0.2 was given.

For simultaneous steering, we  calculate the matching scores for each behaviour while steering with the same vector. 
To measure the alignment tax, we also calculate the top-1 accuracy on 500k tokens from the Pile, containing both text and code. 
We calculate this top-1 accuracy while the model is being steered for each global injection coefficient. 
\section{Results} \label{results}
\subsection{Broad steering} \label{results broad steering}

\begin{figure}
\centering
\begin{subfigure}[t]{0.45\linewidth}
\includegraphics[width=\linewidth]{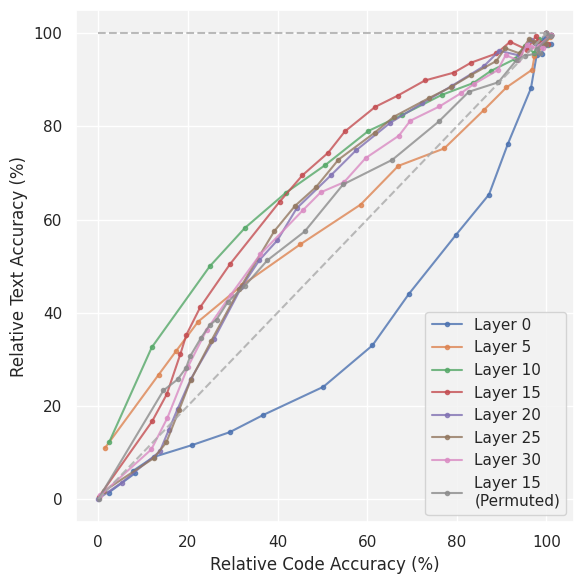}
    \caption{The overall performance.}
    \label{fig:coding ability}
\end{subfigure}
\begin{subfigure}[t]{0.45\linewidth} 
    \includegraphics[width=\linewidth]{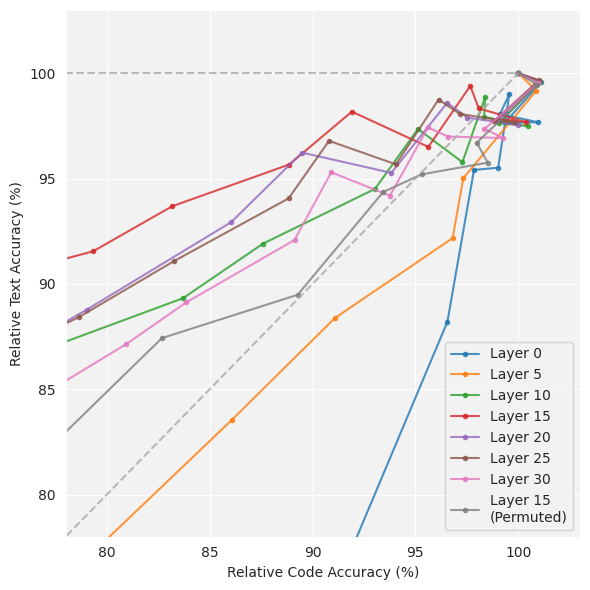}
    \caption{The same as in (a), but cropped to the top scores.}
    \label{fig:coding ability cropped}
\end{subfigure}
\caption{This figure illustrates the effect of applying a steering vector aiming to remove coding ability. The scores for text and code data are recorded. Each line represents steering at a certain layer, and each dot represents the scores for one injection coefficient. The horizontal dotted line illustrates perfect performance (which is impossible for our data due to noise), and the diagonal dotted line shows equal performance drops in code and text.}
\label{fig:general label for coding}
\end{figure}

\begin{figure}
\centering
\begin{subfigure}[t]{0.45\linewidth}
    \includegraphics[width=\linewidth]{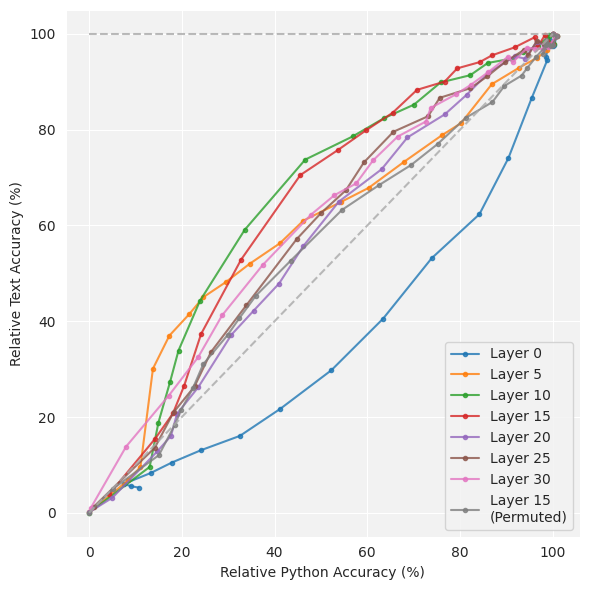}
    \caption{Subfigure Python ability}
    \label{fig:python ability}
\end{subfigure}
\begin{subfigure}[t]{0.45\linewidth}
    \includegraphics[width=\linewidth]{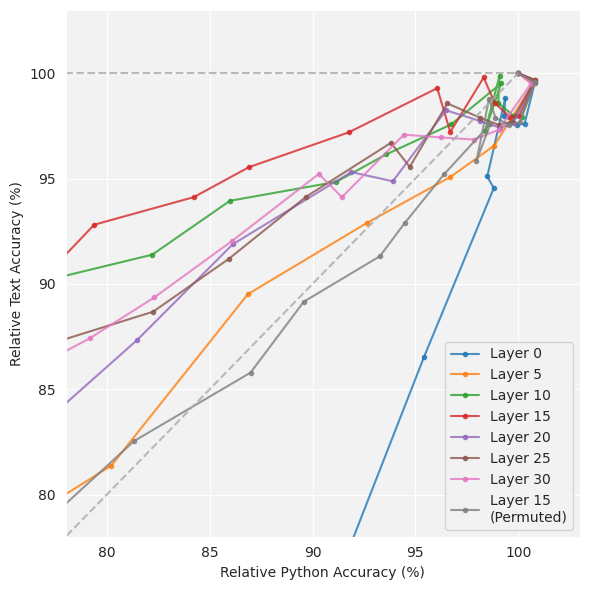}
    \caption{Python ability cropped.}
    \label{fig:python ability cropped}
\end{subfigure}
\caption{This figure illustrates the effect of applying a steering vector aiming to remove Python ability, in the same way as in Figure \ref{fig:general label for coding}.}
\label{fig: general label for python}
\end{figure}

Figure \ref{fig:general label for coding} shows the relative coding vs textual performance for top-1 next token prediction accuracy, and Figure \ref{fig: general label for python} shows this for Python-specific performance. We first describe the results in Figure \ref{fig:general label for coding}, and afterwards compare it with Figure \ref{fig: general label for python}.

After subtracting the coding steering vector and adding the text steering vector for various injection coefficients, we see that activation steering works to relatively reduce coding ability for most layers. For example, a 60\% relative  top-1 coding accuracy (40\% fewer correctly predicted tokens) corresponds to an 80\% relative textual accuracy for layer 15. Because developers will likely only accept a marginal penalty on general performance, Figure \ref{fig:coding ability cropped}, sheds light specifically on these smaller margins. We see a 10\% reduction in coding performance, corresponding to only a 3\% reduction in textual performance. The permuted steering vectors show a pattern similar to worse performing layers. The steering vector for layer 0 surprisingly produced the opposite of the intended effect. 

We hypothesized that steering for broader concepts would work but with a smaller effect size compared to narrower steering. Figure \ref{fig:python ability} shows similar results to Figure \ref{fig:coding ability}, contradicting our hypothesis. In the cropped Figure \ref{fig:python ability cropped}, we again see a similar performance to general coding. Although the steering vectors clearly work, the selective pruning method by \citet{pochinkov2023dissecting} is substantially more effective.

\clearpage

\subsection{Multi-steering} \label{results multi-steering}

\begin{wrapfigure}{r}{0.5\textwidth}
    \vspace{-0.2cm}
    \centering
    \includegraphics[width=\linewidth]{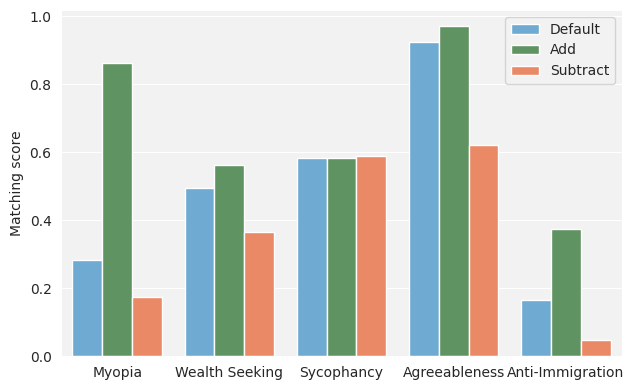}
    \caption{The results for individual steering for layer 15. The used injection coefficients can be found in Table \ref{tab:injection coefficients single steering}}
    \label{fig: single steering layer 15}
    \vspace{-0.2cm}
\end{wrapfigure}
Here we show the results for layer 15, see Appendix \ref{app: layer 10 results} for the results for layer 10. In Figure \ref{fig: single steering layer 15} we see that for myopia, wealth seeking, agreeableness, and anti-immigration individual steering works in both directions but with varying effect sizes. Interestingly, steering for sycophancy has a negligible effect in our experiment, in contrast with previous work \citep{rimsky2023steering}. The difference is likely due to our focus on political topology, and not other forms of sycophancy.

\begin{figure}[h]
  \centering
  \includegraphics[width=\linewidth]{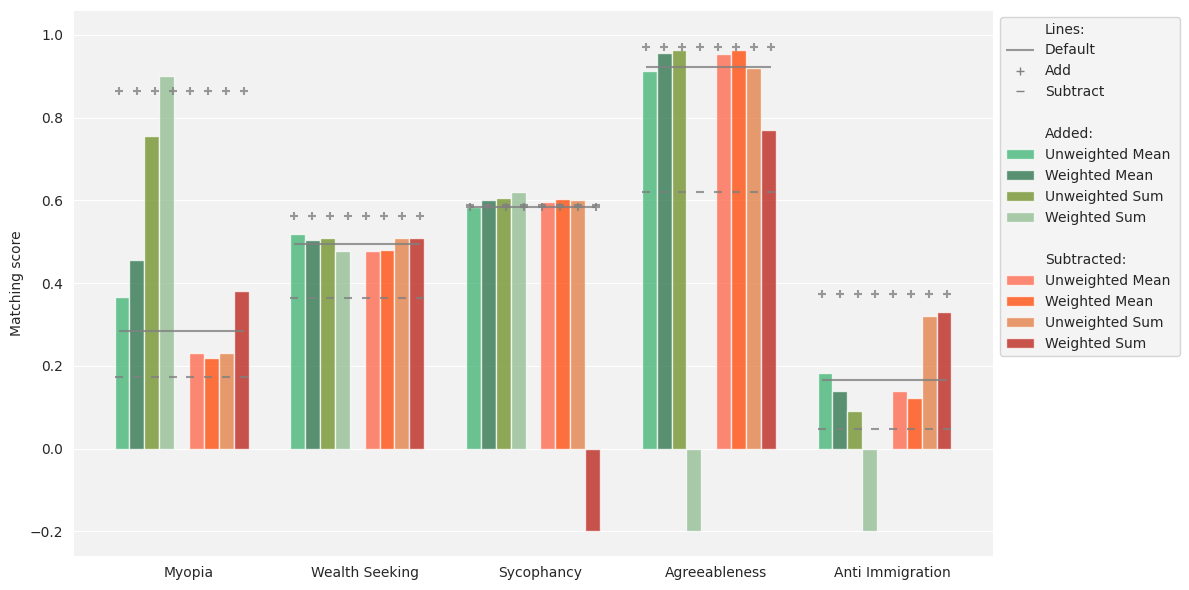}
  \caption{Combining the individual steering vectors into one injected in layer 15. The combinations differ in three dimensions: take the mean or sum, weighted or unweighted, and subtracted or added. We compare the combined steering to the individual steering presented in Figure \ref{fig: single steering layer 15}, which are indicated with the grey horizontal lines.}
  \label{fig: layer 15 multiple steering}
\end{figure}

The results of the 8 combined steering vectors are shown in Figure \ref{fig: layer 15 multiple steering}. We observe that combined steering leads to unexpected and often smaller effect sizes than steering individually. The smaller effect sizes are clearly visible for the first three combinations for addition and subtraction. The last combination for myopia demonstrates unexpected effects: the weighted summation steering vector for addition has a larger effect size than with individual steering, but the subtracting weighted summation for subtraction leads to a higher matching score than without steering at all. Moreover, for anti-immigration, we see that adding generally leads to lower matching scores and subtracting to higher matching scores, which is also in contrast to expectation. For wealth seeking in particular, none of the combined steering vectors maintained a substantial effect. We see that combined steering vector leads to mode collapse in some cases, indicating an increased alignment tax.

\begin{figure}[ht]
  \centering
    \includegraphics[width=\linewidth]{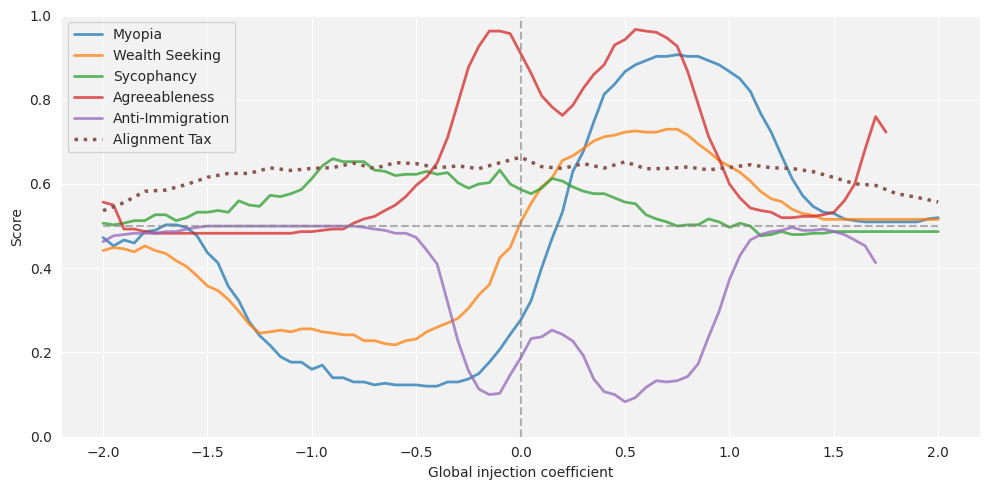}
  \caption{The results for simultaneous steering. Each individual steering vector is injected at a different layer, according to the same global injection coefficient on the x-axis. The score is the matching score for the 5 behaviours, and is the top-1 prediction accuracy score for the alignment tax. A piece of line  missing indicates that there were more than 5\%  faulty responses.}
  \label{fig: simultaneous steering}
\end{figure}

In Figure \ref{fig: simultaneous steering} we show the effect of simultaneous steering at multiple layers of the residual stream at the last token. For myopia and wealth seeking behaviours, the effects are comparable to individual steering. For anti-immigration and agreeableness the effect sizes are smaller and more unstable than with individual steering. For anti-immigration the steering only works for small injection coefficients, and for agreeableness the opposite effect occurs for small injection coefficients. As expected based on individual steering, sycophancy steering does not work here either.
Additionally, as the absolute global injection coefficients increase, we see the effect of mode collapse arising. The scores converge to the dotted brown horizontal line at 0.5, which is the score when each answer has the same token, e.g. `Yes'. The alignment tax appears to be minor; there is only a couple percentage points decrease for global injections coefficients of -1 and +1. 

\section{Discussion} \label{discussion}
\subsection{Broad steering} \label{discussion coding experiments}

In Figure \ref{fig:general label for coding} and Figure \ref{fig: general label for python} we see that activation steering can work for broad skills (coding ability) and is competitive with steering towards narrower skills (Python ability).

One possible explanation for this counter-intuitive result is that the steering vectors for Python and general coding roughly equally distort the model's workings. 
However, the spread of the activation distribution in Appendix \ref{app: activation distribution} seems to provide evidence against this explanation, as the spread of Python activations is smaller than the general coding steering vector for layer 15. 
Steering vectors are also multiplied by the same injection coefficients, so this is not a factor either. 

Another explanation for the result that activation steering works well for broad skills could be that Python data constituted a large part of the overall coding data for Llama 2, and therefore the results are highly correlated. 
This explanation cannot be easily verified or refuted because the training data of Llama 2 is private. Additional experiments removing programming languages other than Python might be informative here.

Furthermore, the opposite effect of steering at layer 0 (where steering to remove coding ability leads to an \emph{improvement} in coding) is confusing. 
One speculative theory comes from a paper introducing the concept of copy suppression, which states that “[i]f components in earlier layers predict a certain token, and this token appears earlier in the context, the attention head suppresses it” \citep{mcdougall2023copy}. 
It could be that coding activations do not occur in earlier layers due to steering, and therefore the copy suppression in subsequent layers does not work. More research is needed to support or refute this theory.

\subsection{Multi-steering} \label{multi-steering experiments}
In line with previous work \citep{rimsky2023steering, turner2023activation, zou2023representation, li2023inference}, we find that steering vectors for individual behaviours are effective. 
Below we discuss our findings for steering multiple behaviours.

\paragraph{Combined Steering} 
In Figure \ref{fig: layer 15 multiple steering} we show that combining these individual steering vectors into one steering vector is less successful: we only find substantial effect sizes in the desired direction for myopia. 
This overall reduced steerability result aligns with our hypothesis. 
However, it is possible that another method for combining steering vectors or other hyperparameter settings lead to a larger and more reliable effect. 
At least, we have illustrated that combined steering is not straightforward. 
In particular, the easiest methods for combining vectors result in ineffective steering vectors. 

\paragraph{Simultaneous Steering}
In Figure \ref{fig: simultaneous steering} we find that simultaneous injection of individual steering vectors at different places in the model appears more effective than combined steering. 
In particular, we find that we can substantially and reliably steer two behaviours (myopia and wealth seeking). 
For agreeableness and anti-immigration behaviours we find a minor and less reliable effect. 
This increased steerability is likely due to a lower disturbance of the activation pattern of each individual behaviour. 
Moreover, our results suggest that interaction effects (between steering at different layers) do not substantially reduce the steering effect. 
Therefore, interaction effects from simultaneous steering appear less problematic than the changed direction in the latent space with combined steering. 
Surprisingly, simultaneous steering merely leads to a marginal alignment tax. 

All in all, simultaneous steering seems like a more promising method than combined steering.

\subsection{General discussion} \label{general discussion}
The flexibility of activation steering is a double-edged sword. 
The large variety of activation generation and injection techniques affords broad applicability, but finding an optimal setup is not straightforward. 
This could be due to the novelty of the method, as best practices have not yet been established. 
This large variety is also mentioned in the original Activation Addition paper \citep{turner2023activation}.
As a result, the claim that activation steering does not work well is not a certain claim; the possibility exists that a slightly different approach would be fruitful.

Moreover, as we care about reducing risks from models, we need to know the `real' performance of models, not what matching score they achieve. 
Related work has illustrated that the matching does translate to open-ended generation \citep{rimsky2023steering}, which indicates that matching score might be a reasonable proxy for `real' performance. 
However, it is still unclear whether e.g. a 20\% reduction in top-1 coding accuracy would actually reduce risks when coding ability of a model is risky. 

\subsection{Future work}
To further investigate broad steering capabilities, more narrow skills can be investigated. 
For example, if Python makes up a large part of the coding data in the Llama 2 training set, 
then steering against alternative programming languages may lead to more interesting effects.

To perform simultaneous steering (see Figure \ref{fig: simultaneous steering}) we only vary one global injection coefficient. 
We expect that our results can be improved by using different injection coefficients for different steering vectors. 
Furthermore, the steering vectors can be generated and injected at different places within a layer, such as in the attention and MLP components.
This might allow us to: 
1) find areas that correspond more specifically to a target skill or behaviour; and 
2) find distinct areas for different target behaviours so that we can use simultaneously steer for even more concepts.

Future work can also focus on testing how these results hold up with other models.
This will provide insight in the general working of activation steering.
It seems plausible that sparse language models might be easier to steer than dense language models such as Llama 2, as the activation patterns might be cleaner. Sparse steering is especially promising when extending activation addition due to a reduction in the clashing of activation patterns in individual steering vectors.
\newpage
\section*{Contributions}
Teun van der Weij was the main author of this paper. Nandi Schoots and Massimo Poesio supervised the project, with Nandi Schoots supervising for longer and more intensively.

\section*{Acknowledgments}
We want to thank numerous people for their research ideas, code samples and feedback in general: Nina Rimsky, Nicky Pochinkov, Alex Jackson, Andrea Bruera, Ole Jorgensen, and Nadja Flechner. We also want to thank the research engineering team from Utrecht University for the high-performance cluster support.
\bibliographystyle{apalike}
\bibliography{references}

\clearpage
\newpage
\appendix  

\section{Detailed methodology} \label{detailed methodology}
Here are some additional details to the methodology sections.
\subsection{Methodology relating to all experiments}

\begin{description}
    \item[Truncation] All samples were truncated to 4096 tokens.  
    \item[Model details] The model's dtype was bfloat16.
    \item[Datasets] All datasets were initially randomly shuffled with seed 13.
    \item[Text generation] All generated text was produced without sampling. 
\end{description}

\subsection{Broad steering experiments} \label{app: broad steering}
We used the following injection coefficients: 
\begin{align*}
&0.0, 0.25, 0.5, 0.6, 0.7, 0.8, 0.9, 1.0, 1.1, 1.2, 1.3, 1.4, 1.5, 1.6, 1.7, 1.8, \\
&1.9, 2.0, 2.25, 2.5, 2.75, 3, 3.5, 4, 4.5, 5, 6, 7, 8, 9, 10, 15, 20, 30, 40, 50
\end{align*}

There are some criteria for which we did not stepwise go through these values. We calculate the steered top-1 accuracy score divided by the default score. Based on this relative score, we used the following strategy to go through the injection coefficients. The score of either general or Python code was below 0.05, the run was stopped. Otherwise, if the relative score was below 0.15, a step size of 5 would be taken instead of 1.

\subsection{Multi-steering experiments}
\begin{description}
    \item[Grid search] Extending on the grid search described in Section \ref{multiple steering activation injection}. The tested injections coefficients values were $\{0.5, 1, 2, 3, 5, 10, 20, 30, 40, 60, 80, 120, 200, 300\}$, and the tested layers were $\{0, 5, 10, 15, 20, 25, 29, 31\}$. The specific injection coefficients per behaviour are shown in Table \ref{tab:injection coefficients single steering} for layers 10 and 15, the two most effective layers.
\end{description}

\begin{table}[ht]

\centering
\begin{tabular}{@{}llllll@{}}
\toprule
         & Agreeableness & Anti Immigration & Myopic & Wealth seeking & Sycophancy \\ \midrule
Layer 10 & 0.5, -3       & 3, -1            & 10, -1 & 1, -2          & 1, -20     \\
Layer 15 & 0.5, -1       & 1, -0.5          & 2, -1  & 1, -2          & 2, -5      \\ \bottomrule
\end{tabular}

\caption{The injection coefficients for each concept after performing grid search for adding and subtracting the steering vectors for layers 10 and 15.}
\label{tab:injection coefficients single steering}
\end{table}

\newpage
\section{Layer 10 results} \label{app: layer 10 results}
\subsection{Single steering}

\begin{figure}[h]
  \centering
    \includegraphics[width=0.5\linewidth]{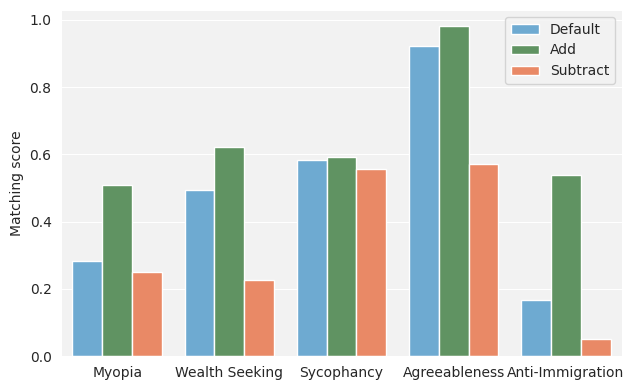}
\caption{The results for individual steering for layer 10. The used injection coefficients can be found in Table \ref{tab:injection coefficients single steering}}
  \label{fig: single steering layer 10}
\end{figure}

\subsection{Combined steering}
\begin{figure}[h]
  \centering
  \includegraphics[width=\linewidth]{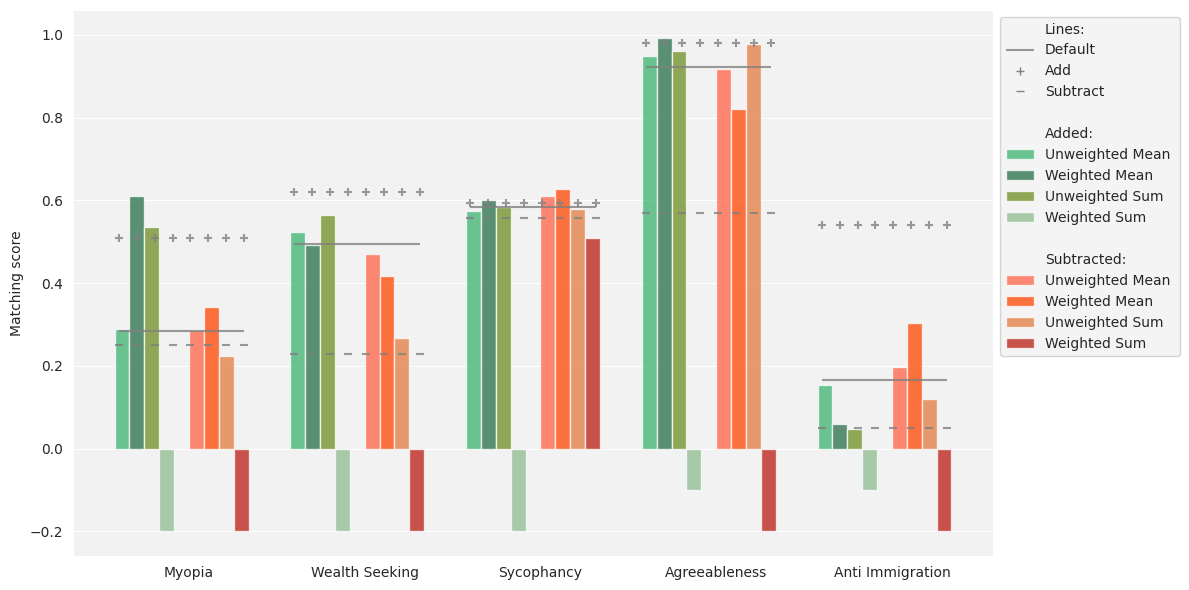}
    \caption{Combining the individual steering vectors into one injected in layer 10. The combinations differ in three dimensions: take the mean or sum, weighted or unweighted, and subtracted or added. We compare the combined steering to the individual steering presented in Figure \ref{fig: single steering layer 10}, which are indicated with the grey horizontal lines.}
  \label{fig: layer 10 multiple steering}
\end{figure}
\clearpage
\newpage

\section{Activation distribution} \label{app: activation distribution}
\subsection{General coding activation distributions}
\begin{figure}[h]
\label{general coding activation distributions}
    \centering
    \includegraphics[width=0.8\linewidth]{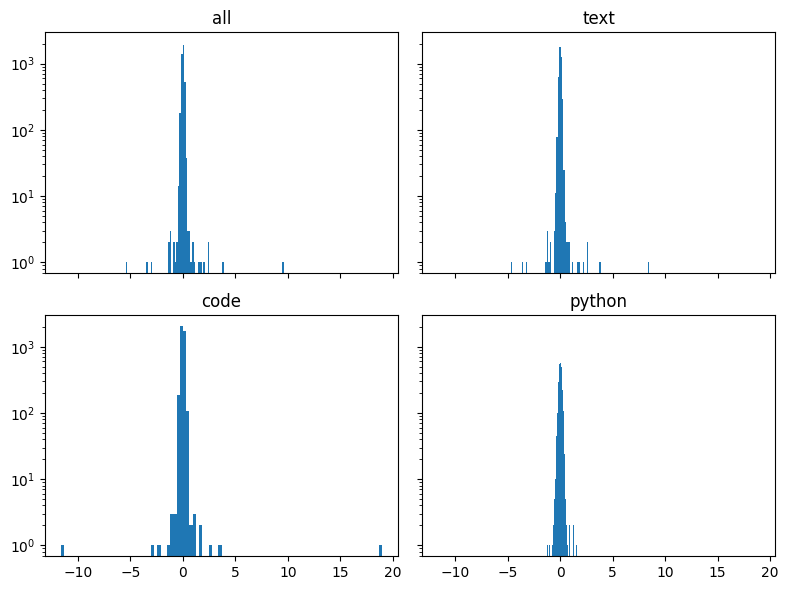}
    \label{fig:activation distribution coding}
    \caption{The distribution of activations for layer 15 given multiple datasets.}
\end{figure}

\subsection{Multi steering activation distributions} \label{multi steering activation distributions}
\begin{figure}[h]
\includegraphics[width=\linewidth]{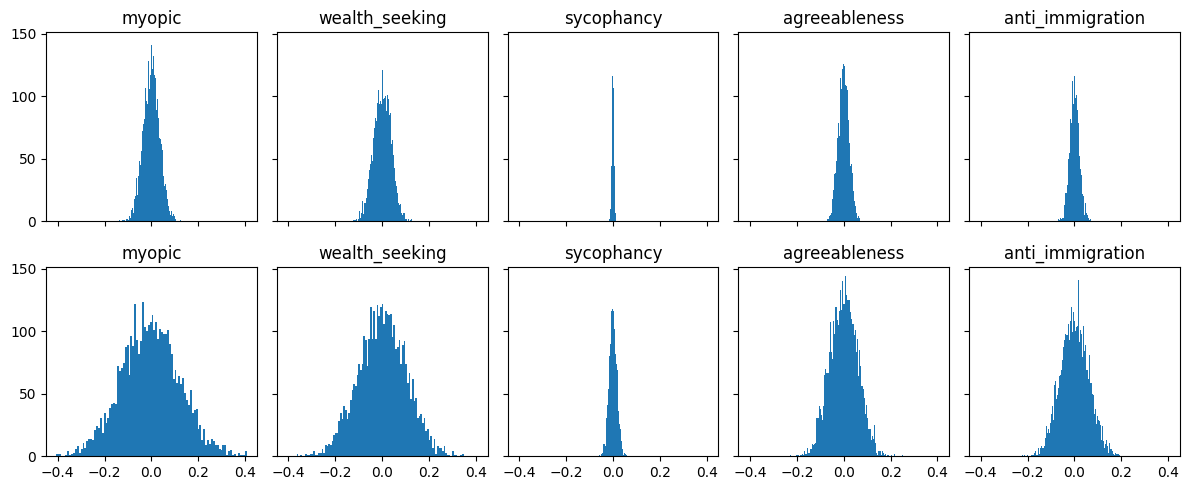}
    \caption{The distribution of activations for layers 10 (above) and 15 (below) for numerous concepts.}
    \label{fig:activation distribution multi steering}
\end{figure}

\end{document}